\documentclass{article}
\usepackage{arxiv}
\usepackage{sistyle}
\SIthousandsep{,}
\usepackage[noend]{algorithmic}

\renewcommand\algorithmicthen{}

\newcommand{\ONELINEIF}[2]{\STATE \algorithmicif\ #1\ \algorithmicthen\ #2}
\usepackage[utf8]{inputenc}
\usepackage{color}
\usepackage{amssymb}  % assumes amsmath package installed
\usepackage{ucs}
\usepackage{comment}
\usepackage{amsmath}
\usepackage{pbox}%fait des boites
\usepackage{tcolorbox}
\usepackage{caption}
\usepackage{capt-of}
\usepackage{bm}
\usepackage{graphicx}

\usepackage{array}
\usepackage{multirow}
\usepackage{colortbl}
\usepackage{tikz}
\usepackage{xspace}
\usepackage[caption=false,font=footnotesize]{subfig}

\usepackage{subfig}

\newcommand{\ddpg}{{\sc ddpg}\xspace}

\newcommand{\cmaes}{{\sc cma-es}\xspace}

\newcommand{\openes}{{\sc open-es}\xspace}
\newcommand{\snes}{{\sc snes}\xspace}
\newcommand{\nes}{{\sc nes}\xspace}
\newcommand{\xnes}{x{\sc nes}\xspace}

\newcommand\cem{\textsc{cem}\xspace}

\newcommand{\picma}{{\sc pi$^2$-cma}\xspace}

\newcommand{\drl}{deep RL\xspace}

\newcommand{\edas}{{\sc EDA}s\xspace}
\newcommand{\eda}{{\sc EDA}\xspace}

\newcommand{\mymath}[1]{\ensuremath{#1}\xspace}
\newcommand{\mymathbf}[1]{\mymath{\mathbf{\boldsymbol{#1}}}}
\newcommand{\covar}{\mymathbf{\Sigma}}

\renewcommand{\vec}[1]{\mymathbf{#1}}

\newcommand\wi[1]{$\circ$}
\newcommand\bu[1]{$\bullet$}
\newcommand\ot[1]{$\star$}
\newcommand\spa[1]{$\spadesuit$}
\newcommand\dn[1]{.}

\newcommand{\params}{{\mymathbf{\theta}}\xspace} % policy parameters (vector)

\newcounter{cbox} 
\newcommand{\cbox}{\arabic{cbox}}

\newcounter{cmes} 
\newcommand{\cmes}{\arabic{cmes}}

\usepackage{algorithmic}
\usepackage{algorithm} 

\makeatletter
\newcounter{algorithmbis}
\renewcommand{\thealgorithmbis}{\thesection.\arabic{algorithmbis}}
\def\algorithmbis{\@ifnextchar[{\@algorithmbisa}{\@algorithmbisb}}
\def\@algorithmbisa[#1]{%
  \refstepcounter{algorithmbis}
  \trivlist
  \leftmargin\z@
  \itemindent\z@
  \labelsep\z@
  \item[\parbox{\linewidth}{%
    \hrule
    \hrule
    \noindent\strut\textbf{Algorithm \thealgorithmbis} #1
    \hrule
  }]\hfil\vskip0em%
}
\def\@algorithmbisb{\@algorithmbisa[]}

\makeatother

\definecolor{myred}{rgb}{0.8,0,0}
\definecolor{mygreen}{rgb}{0,0.6,0}
\definecolor{myblue}{rgb}{0,0,0.7}

\date{}

\newcommand{\vth}{\mymath{\vec \theta}}

\newcounter{rules} 
\newcommand{\rules}{\arabic{rules}}
\newenvironment{myrule}[1]{\vspace{0.3cm}\begin{tcolorbox}[colback=blue!10!white]\refstepcounter{rules}{\bf Rule \rules:}\label{#1}}{\end{tcolorbox}\vspace{0.3cm}}

\newcounter{theorems} 
\newcommand{\theorems}{\arabic{theorems}}
\newenvironment{mytheorem}[1]{\vspace{0.3cm}\begin{tcolorbox}[colback=blue!10!white]\refstepcounter{theorems}{\bf Theorem \theorems:}\label{#1}}{\end{tcolorbox}\vspace{0.3cm}}

\begin{document}
%\begin{frontmatter}
  
\title{Importance mixing: Improving sample reuse in evolutionary policy search methods}
\author{Alo\"{i}s Pourchot, Nicolas Perrin, Olivier Sigaud\\
Sorbonne Universit\'e, CNRS UMR 7222,\\
      Institut des Syst\`emes Intelligents et de Robotique, F-75005 Paris, France\\
      {\tt olivier.sigaud@isir.upmc.fr}~~~~+33 (0) 1 44 27 88 53
}

    \maketitle

  %\begin{keyword}
    %importance mixing, sample efficiency, deep reinforcement learning, deep neuro-evolution
    %Regression Trees
  %\end{keyword}

    \begin{abstract}
Deep neuroevolution, that is evolutionary policy search methods based on deep neural networks, have recently emerged as a competitor to deep reinforcement learning algorithms due to their better parallelization capabilities. However, these methods still suffer from a far worse sample efficiency.
      In this paper we investigate whether a mechanism known as "importance mixing" can significantly improve their sample efficiency. We provide a didactic presentation of importance mixing and we explain how it can be extended to reuse more samples. Then, from an empirical comparison based on a simple benchmark, we show that, though it actually provides better sample efficiency, it is still far from the sample efficiency of deep reinforcement learning, though it is more stable.
    \end{abstract}
%\end{frontmatter}

\section{Introduction}

Policy search is a specific instance of 
black-box optimization where one tries to optimize a policy $\pi_\vth$ parametrized by a vector $\vth$ with respect to some unknown utility function. 
The utility function being unknown, the system must be run several times to get the utility resulting from using various policies $\pi_\vth$ and find a high performing one. Besides, when running the system is costly, it is better to perform as few evaluations as possible, raising a sample efficiency concern \citep{sigaud2018policy}.

Recently, research on policy search methods has witnessed a surge of interest due to the emergence of efficient deep reinforcement learning (RL) techniques which proved stable and sample efficient enough to deal with continuous action domains \citep{lillicrap2015continuous,schulman2015trust,schulman2017proximal}.
In parallel, evolutionary methods, and particularly deep neuroevolution methods applying Evolutionary Strategies (ES) to the parameters of a deep network emerged as a competitive alternative to deep RL due to their higher parallelization capability \citep{salimans2016weight,conti2017improving,such2017deep}.
However, most often these methods are significantly less sample efficient than deep RL methods. One reason for this limitation is that evolutionary methods work with samples consisting of complete episodes, whereas deep RL methods use elementary steps of the system as samples, thus they exploit more information \citep{sigaud2018policy}.

Another reason is that, in general, ES algorithms do not use any memory: the policy $\pi_\vth$ sampled at time step $t$ and its evaluation are discarded when moving to another policy. 
An exception to this second limitation is the "importance mixing" mechanism, which reuses samples from the previous ES generation when building the current generation. This mechanism was already shown to improve sample efficiency in \cite{sun2009efficient} but, to our knowledge, it was never used nor extended in any recent deep neuroevolution method.

Thus one may wonder whether using importance mixing can make evolutionary methods sample efficient enough to compete with deep RL in that respect. In this paper, we address the above question and provide the following contributions.
 
First, in Section~\ref{sec:background}, we give a didactic presentation of the importance mixing mechanism, explaining in particular why not all samples from the previous generation can be reused. Doing so, we also propose an improvement to the importance mixing mechanism initially published in \cite{sun2009efficient}, which consists in successively reusing samples from previous generations until the new generation is full. We describe this improvement in Section~\ref{sec:extension}.
 
Third, in Section~\ref{sec:study}, we investigate experimentally the respective impacts on sample efficiency of the importance mixing mechanism and of the proposed extension, using various ES algorithms. We show that in general, standard importance mixing significantly improves sample efficiency, but that the proposed extension does not do much better. 
We provide a principled explanation for the limited effect of the extension and conclude that sample reuse cannot be much improved with respect to what importance mixing already provides.

Finally, we compare the resulting algorithms to \ddpg, a state-of-the-art deep RL algorithm, on standard benchmarks. The comparison is performed in terms of sample efficiency and shows that importance mixing alone cannot bring enough sample efficiency improvement, though \ddpg is generally less stable.
As a final conclusion, we advocate for different approaches which combine deep RL and evolutionary methods rather than just comparing them.

\section{Understanding Importance Mixing}
\label{sec:background}

In this section, we provide the necessary background to understand the importance mixing algorithm initially published in \cite{sun2009efficient}, first presenting a quick overview of evolutionary strategy algorithms and then giving a detailed presentation of importance mixing itself.

\subsection{Evolutionary Strategies}
\label{sec:es}

Evolutionary algorithms manage a limited population of individuals, and generate new individuals randomly in the vicinity of the previous {\em elite} individuals \citep{back1996evolutionary}.
Evolutionary strategies (ES) can be seen as specific evolutionary algorithms where only one individual is retained from one generation to the next. More specifically, an optimum individual is computed from the previous samples and the next samples are obtained by adding Gaussian noise to the current optimum individual.

Among evolutionary strategies, Estimation of Distribution Algorithms (\edas) are a specific family using a covariance matrix $\covar$ \citep{larranaga2001estimation}.
This covariance matrix defines a multivariate Gaussian function and samples at the next iteration are drawn with a probability proportional to this Gaussian function.
Along iterations, the ellipsoid defined by $\covar$ is progressively adjusted to the top part of the hill corresponding to the local optimum $\params^*$.
The role of $\covar$ is to control exploration. The exploration policy can be characterized as {\em uncorrelated} when it only updates the diagonal of $\covar$ and {\em correlated} when it updates the full $\covar$ \citep{deisenroth2013survey}. The latter is more efficient in small parameter spaces but computationally more demanding and potentially inaccurate in larger spaces as more samples are required. In particular, it cannot be applied in the deep neuroevolution context where the size of \params can be up to millions.

Various instances of \edas, such as \cem, \cmaes, \picma, are covered in \cite{stulp12icml,stulp2012policy,stulp13paladyn}. In this study, we focus on four particular instances of evolutionary methods, listed below. 

\subsubsection{\openes}

First, we consider the ES as implemented in \cite{salimans2017evolution}, which we refer to as \openes. In \openes, the covariance matrix of the Gaussian noise is fixed to $\sigma^2\mathcal{I}_d$, and is not adapted during learning. Only the mean of the distribution is updated through vanilla gradient descent, resulting in a very simple algorithm.

\subsubsection{\snes}

Most ES methods like \openes approximate the vanilla gradient of the utility with respect to the parameters of their sampling distributions.
Instead of doing so, \nes \citep{wierstra2008natural} and \xnes \citep{glasmachers2010exponential} approximate the natural gradient \citep{akimoto2010bidirectional} and use natural gradient descent to update their parameters. Separable Natural Evolution Strategies (\snes), introduced in \cite{Schaul2011HighDA} is a special case of the \nes algorithm where the parameters are uncorrelated, meaning that $\Sigma$ is diagonal. The initial covariance matrix is usually set to $\sigma^2\mathcal{I}_d$,  but unlike in \openes, all diagonal coefficients are adapted during learning. \snes was introduced as a tractable \nes for high dimension optimization problems.

\subsubsection{\cem}

The Cross-Entropy Method (\cem) is a simple \eda where the number of elite individuals is fixed to a certain value $K_e$ (usually set to half the population). After all individuals of a population are evaluated, the $K_e$ most fit individuals are used to compute the new mean and variance of the population, from which the next generation is sampled after adding some limited extra variance to prevent premature convergence.

\subsubsection{\cmaes}

Like \cem, \cmaes is an \eda algorithm where the number of elite individuals is fixed to a certain value $K_e$. The mean and covariance of the new generation are constructed from those individuals. However this construction is more elaborate than in \cem. The top $K_e$ individuals are ranked according to their performance, and are assigned weights conforming to this ranking. Those weights measure the impact that individuals will have on the construction of the new mean and covariance. Quantities called "Evolutionary paths" are also introduced, and are used to accumulate the search directions of successive generations. In fact, the updates in \cmaes are shown to approximate the natural gradient, without explicitly modeling the Fisher information matrix \citep{arnold11informationgeometric}.

\subsection{Rejection sampling}

Consider a random variable $X \in \mathbb{R}^d$ whose probability density function (pdf) $f$ is known. Rejection sampling is a computational mechanism to indirectly sample from $f$. It is based on  the observation that sampling a random variable from its pdf is equivalent to sampling uniformly from the part of the Cartesian space that is below its pdf.

Let us first consider $\mathcal{C}_1$ and $\mathcal{C}_2$, two subsets of $\mathbb{R}^d$ with finite Lebesgue measure. Suppose that $\mathcal{C}_2 \subset \mathcal{C}_1$, and that we know how to sample uniformly from $\mathcal{C}_1$, but not from $\mathcal{C}_2$. In fact, we can sample uniformly from $\mathcal{C}_2$ by trying samples from $\mathcal{C}_1$ until we get samples in $\mathcal{C}_2$, as described in Algorithm \ref{alg:rej_sampling}. Intuitively, if $X$ is a random variable following a uniform law over $\mathcal{C}_1$ (denoted as $X \sim \mathcal{U}(\mathcal{C}_1)$) then samples drawn from $X$ should uniformly recover $\mathcal{C}_1$, hence $\mathcal{C}_2$. Selecting only the samples that landed in $\mathcal{C}_2$ gives the required law, as stated in Theorem \ref{t1}.

\begin{algorithm}[htb]
  \caption{Rejection Sampling}
  \label{alg:rej_sampling}
  \begin{algorithmic}[1]
    \REQUIRE $\mathcal{C}_2 \subset \mathcal{C}_1$ two subsets of $\mathbb{R}^d$ with finite Lebesgue measure, $n$ number of desired samples.
    
    \STATE $Z \leftarrow \emptyset$
    \FOR{$i \leftarrow 1$ to $n$:}
      \STATE Sample $X$ uniformly from $\mathcal{C}_1$
      \WHILE{$X \not\in \mathcal{C}_2$:}
          \STATE  Sample $X$ uniformly from $\mathcal{C}_1$
      \ENDWHILE
      \STATE $Z \leftarrow Z + X$
    \ENDFOR
        
	\RETURN  $Z$
  \end{algorithmic}
\end{algorithm}

\begin{mytheorem}{t1}
Let $\mathcal{C}_2 \subset \mathcal{C}_1$ be two subsets of $\mathbb{R}^d$ with finite Lebesgue measure. Consider a random variable $X \in \mathbb{R}^d$ following a uniform law over $\mathcal{C}_1$. Then the samples $Z=(z_i)_{i=1,\dots,n}$ obtained from Algorithm \ref{alg:rej_sampling} follow a uniform law over $\mathcal{C}_2.$ 
\end{mytheorem}

\begin{figure}[!ht]
  \centering
 \includegraphics[width=0.5\linewidth]{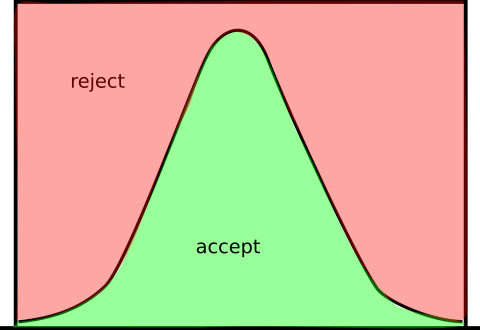}
   \caption{Rejection sampling\label{fig:reject_sampling}}
\end{figure}

A one dimensional (1D) case is illustrated in \figurename~\ref{fig:reject_sampling}. Consider a given rectangle, and a pdf whose support is given by the length of the rectangle. By sampling uniformly from the rectangle and selecting the samples that land below the known pdf $f$, we get a uniform sampling of the area under the curve (in green). 

Going back to the $d$-dimensional case, let us call $\mathcal{C}_f$ the set of points that are under the curve of $f$. Namely, $\mathcal{C}_{f}=\{(x,y)\in \mathbb{R}^d\times\mathbb{R} \textnormal{ s.t. }0\leq y \leq f(x)\}$. The $\mathcal{C}_{f}$ set is the $d$-dimensional equivalent of the area under the curve. Given a pdf $f$, assume we can sample uniformly in $\mathcal{C}_f$. From those samples, could we obtain new samples in $\mathbb{R}^d$  distributed according to $f$? Looking back at the 1D example in \figurename~\ref{fig:reject_sampling}, it would be tempting to select the abscissa of the accepted samples. Indeed, although the accepted points are uniformly distributed in $\mathcal{C}_f$, the abscissa of those points are more likely to fall in regions where the value of the density is high (since there is more vertical space to sample from). Theorem \ref{t2} establishes that this process is indeed probabilistically correct.

\begin{mytheorem}{t2}
Let $X\in \mathbb{R}^d$ and $Y\in \mathbb{R}$ be two random variables, and f a pdf over $\mathbb{R}^d$. If the $(X,Y)$ pair follows a uniform law over $\mathcal{C}_{f}$, then the density of $X$ is $f$.
\end{mytheorem}

Rejection sampling thus solves the problem of sampling directly from $f$ by sampling uniformly from the space under its curve. Finally, if a set $\mathcal{C}$ is known from which we can sample uniformly, then applying Theorem \ref{t1} and \ref{t2} produces samples distributed according to $f$. 

%In the next section, we explain that in the importance mixing mechanism, a good candidate for $\mathcal{C}$ appears effortlessly. Moreover, the candidate found in the importance mixing mechanism can also be parametrized by another density function $g$, from which we can sample directly. 

Finally, we need a way to sample uniformly from the $\mathcal{C}_f$ set, given that we know how to sample from $f$. The 1D example in \figurename~\ref{fig:reject_sampling} can once again help us visualize the process. Sampling from $f$ is sampling an abscissa. Hence, what is left to sample is the corresponding ordinate. A natural idea would be to sample uniformly above the given abscissa $z$, i.e. to get $u$ uniformly sampled on $[0, 1]$ and choose $u.f(z)$ as the ordinate. Theorem \ref{t3} tells us that this is a correct procedure.

\begin{mytheorem}{t3}
Given $X \in \mathbb{R}^d$ a random variable with pdf $f$, $U$ another random variable independent from X following a uniform law over $[0, 1]$. Let $\mathcal{C}_{f}=\{(x,y)\in \mathbb{R}^d\times\mathbb{R} \textnormal{ s.t. }0\leq y \leq f(x)\}$ be the set of points below the graph of $f$. Then the $(X, Uf(X))$ pair follows a uniform law over $\mathcal{C}_{f}$, that is
\begin{align*}
(X, Uf(X)) \sim \mathcal{U}(\mathcal{C}_{f}).
\end{align*}
\end{mytheorem}

%%%%%%%%%%%%%%%%%%%%%%%%%%%%%%%%%%%%%%%%%%%%%%%%%%%%%%
\subsection{Application to importance mixing}

\begin{figure}[!ht]
  \centering
     \includegraphics[width=0.7\linewidth]{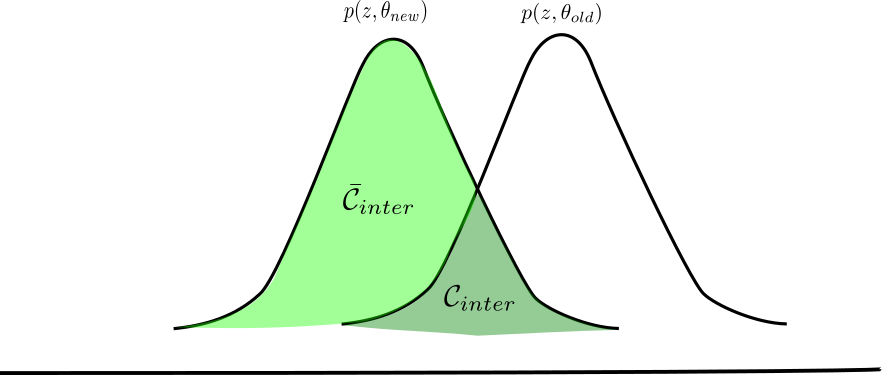}
   \caption{Importance mixing using the previous generation\label{fig:imp_mixing}}
\end{figure}

We now consider two generations $g_{old}$ and $g_{new}$ of any ES described in Section~\ref{sec:es}. Generation $g_{old}$ has been generated by the pdf $p(.,\vth_{old})$ and is of size $N$.
We want to generate Generation $g_{new}$ according to the pdf $p(.,\vth_{new})$ and we also want it to be of size $N$. Moreover, we would like to reuse samples from generation $g_{old}$ into $g_{new}$ without biasing the samples of $g_{new}$ with respect to its pdf. 

Taking a look at \figurename~\ref{fig:imp_mixing}, we can get the intuition of the importance mixing mechanism as follows: to sample from the pdf $p(.,\vth_{new})$, it is enough to uniformly sample from the area below its graph. This area can be split in two parts: the one that intersects with the area below the curve of the old pdf $p(.,\vth_{old})$, $\mathcal{C}_{inter}$, and its complement, $\bar{\mathcal{C}}_{inter}$. Since we already have access to samples from the pdf $p(.,\vth_{old})$, we can sample uniformly below the curve of $p(.,\vth_{old})$ and then uniformly sample from $\mathcal{C}_{inter}$. 

We now need to further detail two sampling rules: one to sample uniformly from the common area $\mathcal{C}_{inter}$, and one to sample uniformly from the leftover area $\bar{\mathcal{C}}_{inter}$. Once this is done, we explain how to combine them.

\subsubsection{Getting samples from the previous generation}

As we stated above, to sample from the common area, we reuse samples coming from generation $g_{old}$. According to Theorem \ref{t3}, to sample uniformly from the area below the curve of the pdf $p(.,\vth_{old})$, we only need to consider an abscissa $z$, sampled according to $p(.,\vth_{old})$, and an ordinate, obtained as $u.p(z, \vth_{old})$ where $u$ is uniformly sampled between 0 and 1. The intersection area can be formally described as
\begin{align}
\mathcal{C}_{inter} = \{(z,y) \in \mathbb{R}^d\times\mathbb{R} \text{ s.t. } 0\leq y \leq \min(p(z,\vth_{old}), p(z,\vth_{new}))\}.
\end{align}
Since $\mathcal{C}_{inter} \subset \mathcal{C}_{p(.,\vth_{old})}$, according to Theorem \ref{t1}, to sample uniformly on $\mathcal{C}_{inter}$ one only needs to take the samples generated uniformly from $\mathcal{C}_{p(.,\vth_{old})}$ that landed in the intersection. In practice, this means that, assuming $p(z,\vth_{old}) >0$, we accept $z$ in the new generation if
\begin{align}
u.p(z,\vth_{old}) &< min(p(z,\vth_{old}), p(z,\vth_{new})), \\
\text{i.e.\ } u & < min(1, \frac{p(z,\vth_{new})}{p(z,\vth_{old})}).
\end{align}
This makes the first acceptance rule:

\begin{myrule}{r1}
Accept $z$ from $g_{old}$ with probability $min(1,\frac{p(z,\vth_{new})}{p(z,\vth_{old})})$.
\end{myrule}

\subsubsection{Sampling from the new generation}

To sample uniformly from the leftover area, we follow a similar process. Since $\bar{\mathcal{C}}_{inter} \subset \mathcal{C}_{p(.,\vth_{new})}$ and $\bar{\mathcal{C}}_{inter} \cap \mathcal{C}_{p(.,\vth_{old})}=\emptyset$, we cannot reuse any more samples. Thus we must sample directly from $p(.,\vth_{new})$. We can describe the leftover region formally as
\begin{align}
\bar{\mathcal{C}}_{inter} = \{(z,y) \in \mathbb{R}^d\times\mathbb{R} \text{ s.t. } \min(p(z,\vth_{old}), p(z,\vth_{new})) \leq y \leq p(z,\vth_{new})) \}.
\end{align}

After sampling $z$ from $p(., \vth_{new})$, given $u$ uniformly sampled between 0 and 1 and assuming $p(z,\vth_{new}) >0$,  we accept $z$ in the new generation if
\begin{align}
u.p(z, \vth_{new}) &\ge \min(p(z,\vth_{old}), p(z,\vth_{new})), \\
\text{i.e.\ \quad} 1-u &\le \max(0, 1 - \frac{p(z,\vth_{old})}{p(z,\vth_{new})}).
\end{align}

This results in the second acceptance rule:
\begin{myrule}{r2}
Draw $z$ as an individual generated from $p(.,\vth_{new})$ and accept it with probability $max(0,1-\frac{p(z,\vth_{old})}{p(z,\vth_{new})})$.
\end{myrule}

\subsubsection{Combining both sources of samples}

Now assume that the $d$-dimensional volume of the intersection is equal to $\lambda$. Since $p(., \vth_{new})$ is a pdf, the volume of the leftover area must be $1-\lambda$. If we want to sample uniformly from $\mathcal{C}_{p(.,\vth_{new})}=\mathcal{C}_{inter} \cup \bar{\mathcal{C}}_{inter}$, we need to sample uniformly from $\mathcal{C}_{inter}$ and $\bar{\mathcal{C}}_{inter}$ according to the ratio of their volume. Fortunately, when sampling uniformly from the volume below the curve of $p(., \vth_{old})$, samples land in the intersection with probability $\lambda$. Likewise, when sampling uniformly from the volume below the curve of $p(., \vth_{new})$, samples land in the leftover volume with probability $1-\lambda$. This means that to sample uniformly on $\mathcal{C}_{p(.,\vth_{new})}$, it is enough to simply use alternatively Rule~\ref{r1} and Rule~\ref{r2}. The complete algorithm is described in Algorithm~\ref{alg:imp_mix}.

\begin{algorithm}[htb]
  \caption{Importance mixing}
  \label{alg:imp_mix}
  \begin{algorithmic}[1]
    \REQUIRE $p(z,\vth_{new})$: new pdf, $p(z,\vth_{old})$: old pdf, $g_{old}$: old generation
    \STATE  $g_{new} \leftarrow \emptyset$
    \FOR{$i \leftarrow 1$ to $N$}
    	\item[]
        \STATE Draw rand1 and rand2 uniformly from $[0,1]$
    	\STATE  $z_i = g_{old}(i)$ \hspace{1.2cm} \COMMENT{taken uniformly from $g_{old}$}
    	\IF {$min(1,\frac{p(z_i,\vth_{new})} {p(z_i,\vth_{old})})>$ rand1:}
    		\STATE  $g_{new} \leftarrow g_{new}.append(z_i)$  \COMMENT{Rule~\ref{r1}}
    	\ENDIF
    	\item[]
    	\STATE  draw $z'_i \sim p(.,\vth_{new})$
        \IF {$max(0,1-\frac{p(z'_i,\vth_{old})}{p(z'_i,\vth_{new})})>$ rand2:}  
    		\STATE  $g_{new} \leftarrow g_{new}.append(z'_i)$  \COMMENT{Rule~\ref{r2}}
   		\ENDIF
    	\item[]
		\STATE  size = $|g_{new}|$
		\ONELINEIF{size $\geq N$:}{go to 12}
    	\item[]
    \ENDFOR
    \ONELINEIF{size $> N$:}{remove a randomly chosen sample}
    \ONELINEIF{size $< N$:}{fill the generation sampling from $p(.,\vth_{new})$}
    \RETURN  $g_{new}$
  \end{algorithmic}
\end{algorithm}

At each step of the algorithm, one alternatively tries an existing sample from $g_{old}$ and a sample drawn from $p(z,\vth_{new})$. Both samples can be either accepted or rejected according to Rules \ref{r1} and \ref{r2} respectively.

Given that we are sampling from two sources at each step, it may happen that, when Generation $g_{new}$ already contains $N-1$ samples, two samples are accepted at the last iteration and we get a generation with $N+1$ samples. In that case, it is necessary to remove one sample, which can be randomly chosen.

If one applies the above process $N$ times then, on average, $(1-\lambda)N$ samples will be accepted from $g_{old}$ and $\lambda N$ samples will be accepted from $p(z,\vth_{new})$, thus on average the size of the obtained generation $g_{new}$ will be $N$.  In practice however, samples may be accepted faster or more slowly than on average. If they are accepted faster, this means that Generation $g_{new}$ will contain $N$ samples {\em before} the above process has been run $N$ times. If they are accepted more slowly, this means that Generation $g_{new}$ will contain less than $N$ samples after the above process has been run $N$ times (which means that all samples from $g_{old}$ have been tried).

In that context, \cite{sun2009efficient} propose instead to first exhaust the old generation using Rule~\ref{r1}, resulting in $N_{\beta} \leq N$ samples. Then, they use Rule~\ref{r2} repeatedly until they get 
$N - N_{\beta}$ new samples. On average, $N$ trials are needed, but it is often the case that more than $N$ or less than $N$ trials are performed, and in these common cases a bias is introduced, as the samples in $\mathcal{C}_{inter}$ and $\bar{\mathcal{C}}_{inter}$ are not well balanced.

%accept exactly one sample per iteration. To do so, they first try a sample from $g_{old}$, accepting it according to . If it is accepted, they proceed to the next iteration. Otherwise, they sample from $p(., \vth_{new})$ and accept the result according to , doing so until a sample gets accepted. By following this process, exactly $N$ samples are added in Generation $g_{new}$. However, this process introduces some bias, as samples from $g_{old}$ would generally be under-represented. 

A more satisfactory option consists in completing Generation $g_{new}$ with samples drawn directly from $p(z,\vth_{new})$ (accepting them all) until it is full. Another option consists in applying the correct importance mixing process that we have outlined above, but using other existing generations, as proposed in Section~\ref{sec:extension}.

Note that \cite{sun2009efficient} also use a "minimal refresh rate" parameter $\alpha$ which is set to 0 in our description here, as a different value may only deteriorate sample reuse.

\subsubsection{Complexity of importance mixing}

The complexity of the algorithm can be deduced as follows: in the case where the pdfs are multivariate Gaussian functions, accepting a sample $z$ with probability $\frac{p(z,\vth_{new})}{p(z,\vth_{old})}$ can be written
$$ random < \sqrt{\frac{det(\Sigma_{old})}{det(\Sigma_{new})}}e^{\frac{1}{2}(\mu_{new}-z)^T \Sigma_{new}^{-1}(\mu_{new}-z) - \frac{1}{2}(\mu_{old}-z)^T \Sigma_{old}^{-1}(\mu_{old}-z)}.$$

Computing the ratio thus involves twice the evaluation of the same determinant, and twice the same matrix inversion, as well as four matrix-vector products, so the global complexity is in $O(4d^{2.373})+Nd^2)$. 

%For square matrices $\Sigma$ of size $d$, the number of operations scales in $O(d^3)$ (although it is possible to get this complexity down to $O(d^{2.373}))$. Since we evaluate on average $2N$ ratios, this makes for a total average complexity of $\mathcal{O}(Nd^3)$. 

One can see that the importance mixing mechanism scales poorly with the dimension of the optimization problem. However, in the case where only diagonal covariance matrices are considered, the complexity can be scaled down to $\mathcal{O}(Nd)$, which is much more affordable in practice.

\subsection{Impact on sampling efficiency}
\label{sec:efficiency}

In terms of sample efficiency, the main point about using importance mixing is that only the new samples drawn from $p(.,\vth_{new})$ need to be evaluated by running the system, as the value of samples from $g_{old}$ is already known. Thus, approximately, only $\lambda N$ samples are evaluated instead of $N$ at each generation. However, depending on the ES algorithm, some of the samples from $g_{old}$ might not effectively be reused. In \openes and \snes, all samples are reused to compute the approximation of the gradients. By contrast, in CEM and CMAES, only the $K_e$ elite individuals of the population are used in practice. One may think that this phenomenon somehow hampers sample reuse. However, whatever the fate of samples from $g_{old}$, all samples from $g_{new}$ need to be evaluated, thus ignoring samples from $g_{old}$ has no impact on sample efficiency.

\section{Extending Importance Mixing to multiple generations}
\label{sec:extension}

\begin{figure}[!ht]
  \centering
     \includegraphics[width=0.7\linewidth]{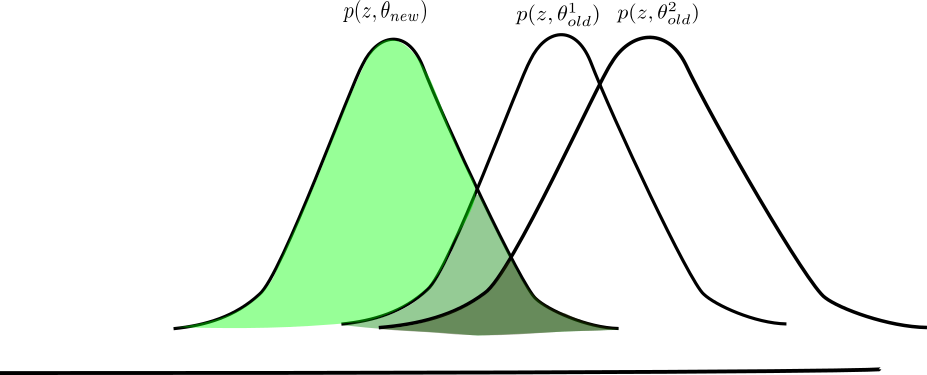}
   \caption{Extended Importance mixing, using several generations\label{fig:ext_imp_mixing}}
\end{figure}

As outlined above, it may happen (roughly half of the time) that after using samples from the previous generation, the new generation is still not full. In that case, samples from former generations could be used to continue filling it while using as few samples from $p(z,\vth_{new})$ as possible, as illustrated in \figurename~\ref{fig:ext_imp_mixing}.

This alternative option is described in Algorithm~\ref{alg:ext_imp_mix}.

\begin{algorithm}[htb]
  \caption{Extended Importance mixing}
  \label{alg:ext_imp_mix}
  \begin{algorithmic}[1]
  \REQUIRE $p(z,\vth_{new})$: new pdf, K: number of generations considered in the archive
  \REQUIRE for $k \in \{1,\dots,K\}$ $p(z,\vth^k_{old})$: $k^{th}$ old pdf, $g^k_{old}$: $k^{th}$ old generation
  \STATE  $g_{new} \leftarrow \emptyset$
  \FOR{$k \leftarrow 1$ to $K$}
		\FOR{$i \leftarrow 1$ to $N$}
    		\item[]
            \STATE Draw rand1 and rand2 uniformly from $[0,1]$
    		\STATE  $z_i = g^k_{old}(i)$ \hspace{1.2cm} \COMMENT{taken uniformly from $g^k_{old}$}
    		\IF {$\min(1,\frac{p(z_i,\vth_{new})}{p(z_i,\vth^k_{old})})>$ rand1:}  
    			\STATE  $g_{new} \leftarrow g_{new}.append(z_i)$  \COMMENT{Rule~\ref{r1}}
    		\ENDIF
    		\item[]
    		\STATE  draw $z'_i$ from $p(.,\vth_{new})$
        	\IF {$\max(0,1-\frac{p(z'_i,\vth^k_{old})}{p(z'_i,\vth_{new})})>$ rand2:}  
    			\STATE  $g_{new} \leftarrow g_{new}.append(z'_i)$  \COMMENT{Rule~\ref{r2}}
    		\ENDIF
    		\item[]
			\STATE  size = $|g_{new}|$
			\ONELINEIF{size $\geq N$:}{{\bf go to} 13}
    	\ENDFOR
	\ENDFOR
    \item[]
    \ONELINEIF{size  $> N$:}{remove a randomly chosen sample}
    \RETURN  $g_{new}$
  \end{algorithmic}
\end{algorithm}

%%%%%%%%%%%%%%%%%%%%%%%%%%%%%%%%%%%%%%%%%%%%%%%%%%%%%%%%%%%%%%%%%%%%%%%%%%%%%%%%%%%
\section{Experimental study}
\label{sec:study}

In this section we investigate the effect of the importance mixing mechanism and its extended version on sample efficiency. We use the \openes,  \snes, \cem and \cmaes evolutionary algorithms presented in Section~\ref{sec:es}.

\subsection{Experimental setup}

To evaluate sample efficiency, we use the CartPole and Acrobot environments from the OpenAI gym benchmark.

In CartPole, a pole is attached to a cart, which is allowed to move along the x-axis. The state information contains the position and velocity of the cart, the velocity of the tip of the pole, and the angle between the pole and the vertical line. At each time step, the agent chooses to push the cart to the left or to the right, with a force of respectively -1 and +1 unit. It receives of reward of +1 at each step, until the episode ends, when either the angle between the pole and the vertical line exceeds 15 degrees, the cart moves more than 2.4 units away from the center, or 200 time steps have elapsed. The maximum possible reward over an episode is thus 200.

In Acrobot, the agent must control a double pendulum (two joints and two links) allowed to move in the xy-plane. Initially, the links are hanging downwards. The state information contains the sine and cosine of the two rotational joint angles and the angular velocities of the two links. At each time step, the agent applies a torque of either -1, 0 or +1 to the joint between the two links. It receives a reward of -1 at each time step, until the episode ends, when either the tip of the second link gets higher than one link above the center joint, or 200 time steps have elapsed. The maximum possible reward in this environment is not known.

Policies $\pi_{\vth_{}}$ are represented by a small neural network with 2 hidden layers of size 8, making for a total of respectively 130 and 155 parameters to be tuned by the evolutionary algorithms for the CartPole and Acrobot environments.

The size of the ES populations is set to 50 and, for all algorithms, the initial covariance is set to $\sigma^2\mathcal{I}_d$, where $\sigma=0.25$. For the algorithms relying on gradient descent (\openes and \snes), we use Adam with an initial learning rate of 0.01, and set the hyper-parameters $\beta_1$ and $\beta_2$ (see \cite{kingma2014adam} for more details on the algorithm) to $0.99$ and $0.999$ respectively. 

We also use a ranking mechanism: instead of raw fitness, we associate to the individuals their rank in the population (fitness-wise). This means that the ES algorithm aims at having individuals with good ranks (and thus good fitness), and it has the advantage to remove the influence of outliers in the population \citep{stulp12icml}. 
Finally, we add weight decay for regularization in all algorithms, which translates into a negative penalty to prevent the weights of the neural networks from being too large. This penalty is computed as $-0.05||\theta||^2$, where $\theta$ is the corresponding set of parameters.

We let each algorithm run for a fixed number of generations: 1000 for \openes and \snes, and 400 for \cmaes and \cem. We do this once with the EIM included, and once without it, and repeat the process for both environments. Results are averaged over 25 runs with different random seeds. The displayed variance corresponds to the $68\%$ confidence interval for the estimation of the mean.

\subsection{Results}
\label{sec:results}

\begin{figure}[!ht]
  \centering
  \subfloat[\label{fig:convergence_cartpole}]{\includegraphics[width=0.49\linewidth]{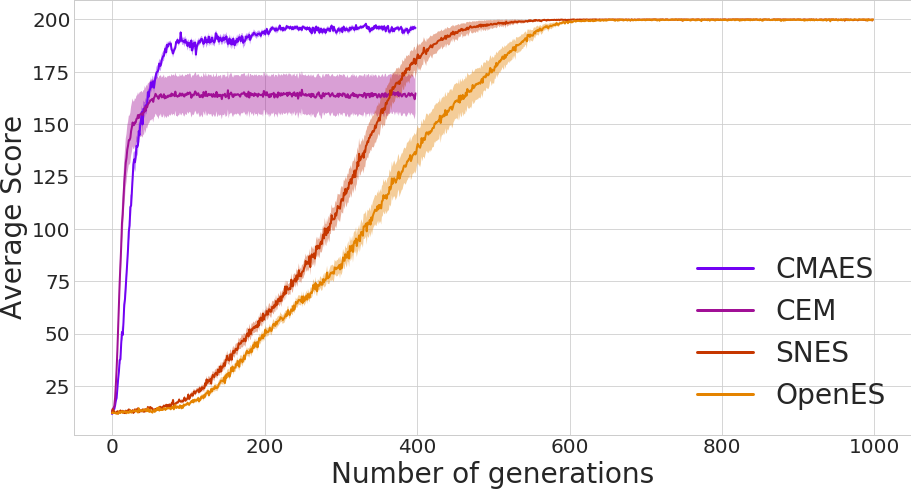}}
     \subfloat[\label{fig:convergence_cartpole_im}]{\includegraphics[width=0.49\linewidth]{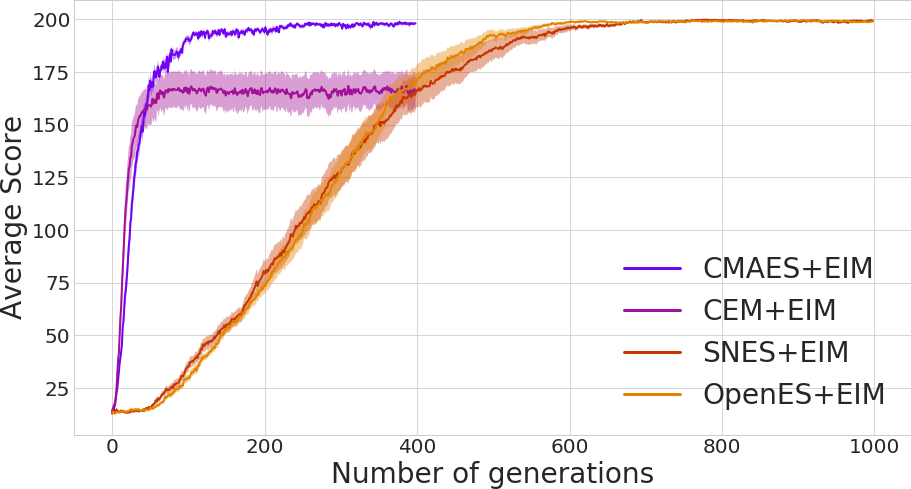}}
   \caption{Evolution of the average score of the population over iterations of the algorithms without (a) and with (b) the extended importance mixing mechanism on CartPole.}
\end{figure}

\begin{figure}[!ht]
  \centering
  \subfloat[\label{fig:convergence_acrobot}]{\includegraphics[width=0.49\linewidth]{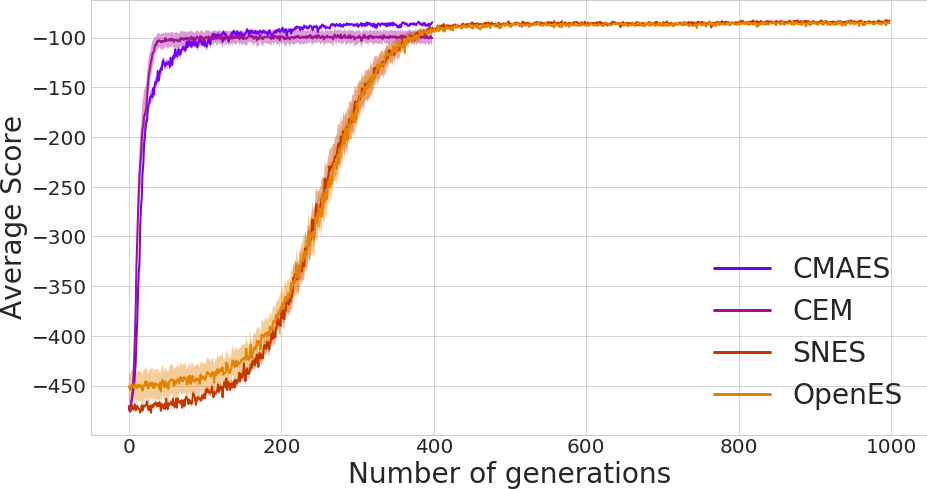}}
     \subfloat[\label{fig:convergence_acrobot_im}]{\includegraphics[width=0.49\linewidth]{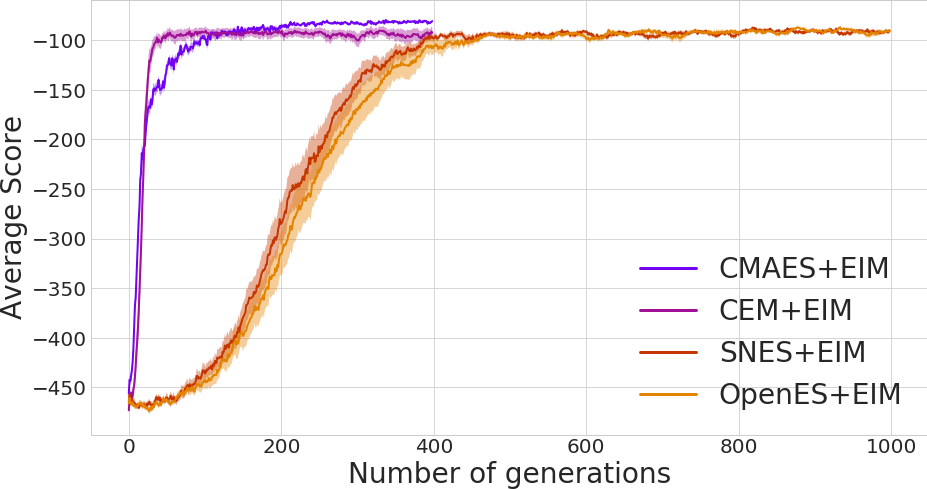}}
   \caption{Evolution of the average score of the population over iterations of the algorithms without (a) and with (b) the extended importance mixing mechanism, on Acrobot.}
\end{figure}

The evolution of the average performance of the population of each algorithm can be seen in \figurename~\ref{fig:convergence_cartpole}, \ref{fig:convergence_cartpole_im} for the CartPole environment, and \figurename~\ref{fig:convergence_acrobot}, \ref{fig:convergence_acrobot_im} for the Acrobot environment. One can see that in general, algorithms have no difficulties finding good policies, but on the CartPole task, \cem seems to get stuck into local maxima.

On CartPole, without the EIM mechanism, the maximum possible reward of $200$ is reached in around $500$ generations with \snes and $600$ with \openes. Interestingly, \cmaes only reaches an average performance around $195$ after $400$ generations whereas \cem gets stuck around $160$ after $100$ generations. With the EIM mechanism, results are very similar, indicating that EIM does not hurt convergence. On the Acrobot environment, without the EIM  mechanism, rewards of around $-90$ are reached in $400$ generations with \snes and \openes, and $50$ generations with \cem. \cmaes slightly outperforms the other algorithms, reaching a fitness around $-75$ in $200$ generations.  Again, the impact of the EIM mechanism on the convergence of the algorithms is limited.

\begin{figure}[!ht]
  \centering
  \subfloat[\label{fig:efficiency_cartpole}]{\includegraphics[width=0.49\linewidth]{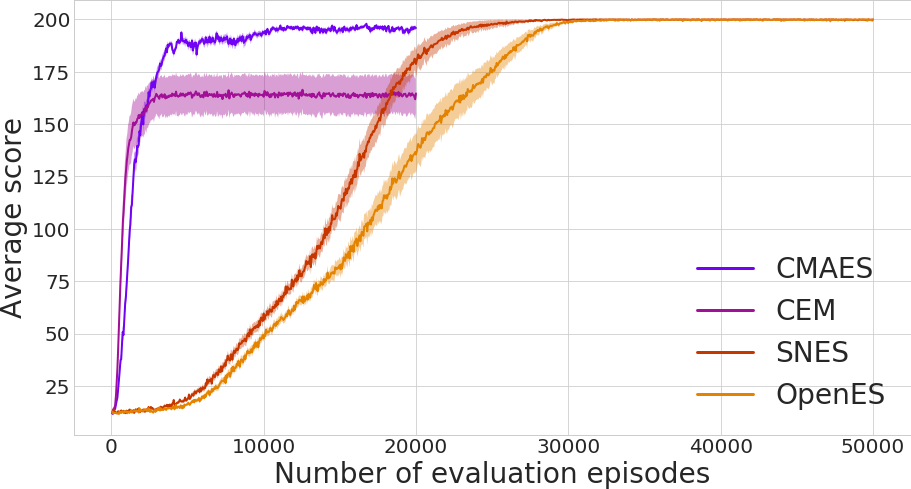}}
     \subfloat[\label{fig:efficiency_cartpole_im}]{\includegraphics[width=0.49\linewidth]{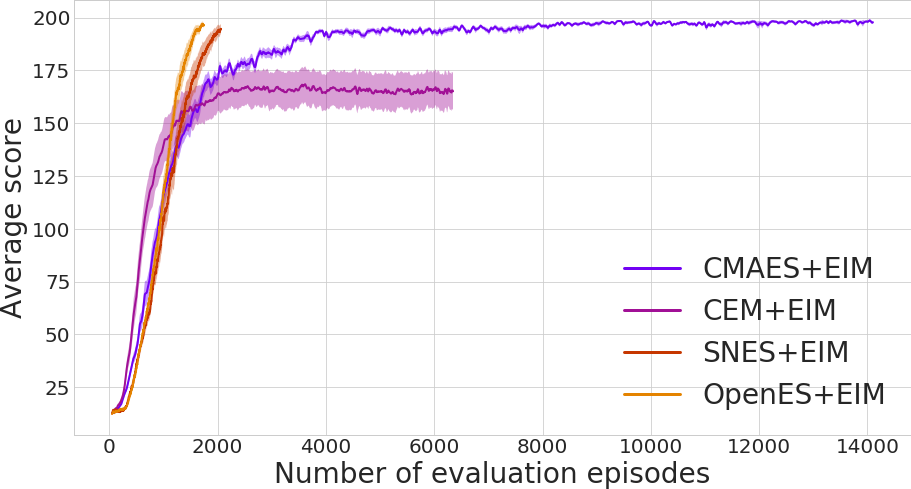}}
   \caption{Evolution of the average score of the population over the number of evaluated episodes without (a) and with (b) the extended importance mixing mechanism on CartPole.\label{fig:cp}}
\end{figure}

\begin{figure}[!ht]
  \centering
  \subfloat[\label{fig:efficiency_acrobot}]{\includegraphics[width=0.49\linewidth]{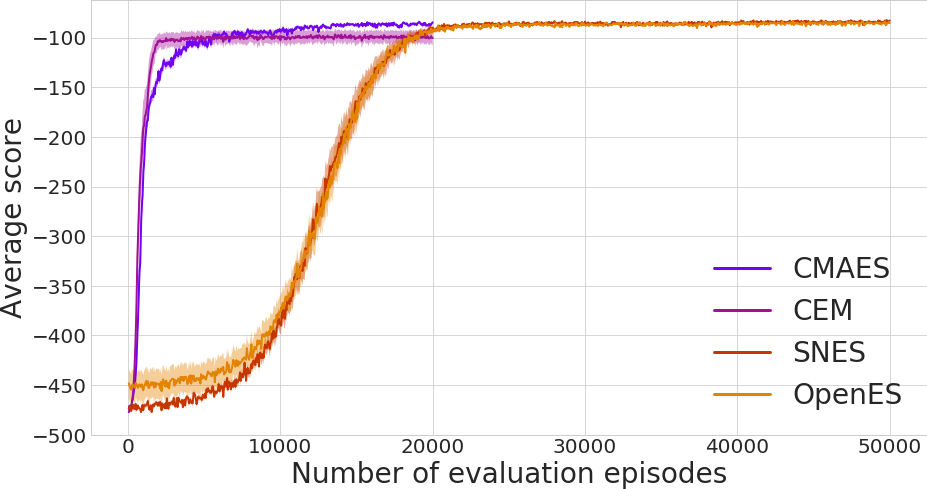}}
     \subfloat[\label{fig:efficiency_acrobot_im}]{\includegraphics[width=0.49\linewidth]{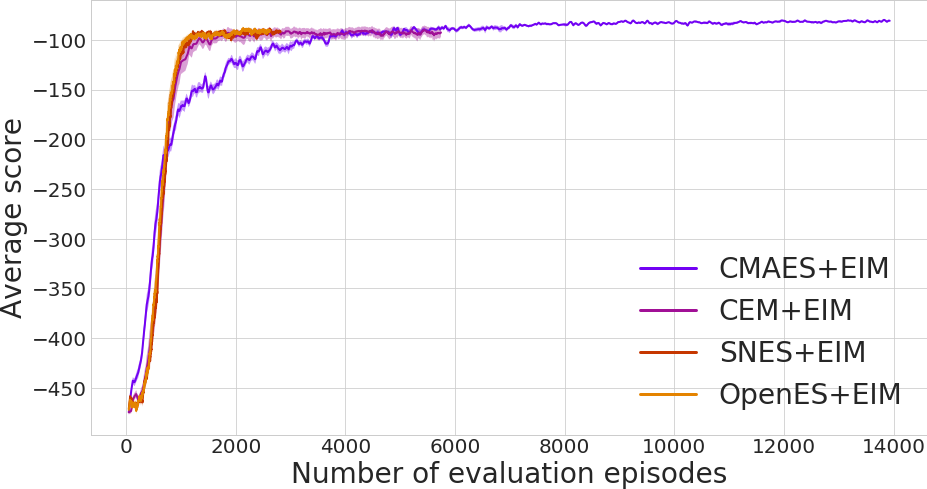}}
   \caption{Evolution of the average score of the population over the number of evaluated episodes without (a) and with (b)  the extended importance mixing mechanism on Acrobot.\label{fig:acro}}
\end{figure}

However, the key quantity in  this study is the number of sample evaluations required to achieve convergence. In \ref{fig:efficiency_cartpole}, \ref{fig:efficiency_cartpole_im}, and \figurename~\ref{fig:efficiency_acrobot}, \ref{fig:efficiency_acrobot_im}, we report the evolution of the average score of the population for all algorithms and both benchmarks, using as time scale the number of samples evaluated from the system, rather than the number of sampled populations.

One can see that the EIM mechanism allows for a substantial speedup in convergence. On CartPole, the total number of evaluations goes down from \num{20000} to \num{14000} for \cmaes, \num{20000} to \num{6000} for \cem and from \num{50000} to $2000$ for \snes and \openes. However, as detailed below, this speedup mostly comes from the IM rather than from its extension.

In Table~\ref{table:sample_reuse_im}, we report the percentage of evaluations that were reused from the previous generation using the importance mixing mechanism and the percentage of evaluations that were reused from the generation preceding the previous generation using the extended importance mixing mechanism. To compute those statistics, we monitored over runs the number of samples reused per generation, and the number of reused samples that came from the EIM.

\begin{table}[ht]
\centering
\begin{tabular}{|c|c|c|c|c|c|c|}
\hline 
	 & \multicolumn{3}{c|}{CartPole Environment} & \multicolumn{3}{c|}{Acrobot Environment} \\
   		 \cline{2-7} 
         & Total reuse & from IM & from EIM & Total reuse & from IM & from EIM \\
  \hline
  \cmaes & $29.2\%$ & $100\%$ & $0.0\%$ & $29.4\%$ & $100\%$ & $0.0\%$ \\
  \hline
  \cem & $50.6\%$ & $90.0\%$ & $10.0\%$ & $48.0\%$ & $89.6\%$ & $10.4\%$ \\
  \hline
  \snes & $95.2\%$ & $96.6\%$ &$3.4\%$ & $93.4\%$ & $96.8\%$ &$3.2\%$ \\
  \hline
  \openes & $95.9\%$ & $97.2\%$ & $2.8\%$ & $94.4\%$ & $97.2\%$ & $2.8\%$  \\
  \hline

\end{tabular}
\caption{\label{table:sample_reuse_im} Sample reuse using importance mixing (IM) and extended importance mixing (EIM) on the CartPole and Acrobot environments. Reuse percentages from IM and from EIM are subparts of the total reuse.} 
\end{table}

From Table~\ref{table:sample_reuse_im}, one can see that in both tasks, \openes and \snes reach excellent reuse rates over $95\%$, meaning that less than one out of ten samples has to be evaluated. On both benchmarks, \cmaes and \cem lag behind with around $30\%$ and $50\%$ sample reuse, which is still non-negligible. Notice that in any case, only around $5\%$ of the reused samples come from the EIM. This quantity is at its maximum and reaches $10.0\%$ and $10.4\%$ for the \cem algorithm, and at its lowest with \cmaes where it is stuck at $0\%$. The boost in sample reuse is thus almost imperceptible.

In this setting, the high reuse rates make \openes and \snes particularly efficient in terms of the number of evaluations, giving them convergence speeds on par with \cmaes and \cem. Their linear complexity also makes them much more affordable in high-dimension.

The fact that the extended importance mixing mechanism does not bring significant improvement over standard importance mixing can be explained easily. On average, standard importance mixing completely fills the new generation half of the time. When the new generation is not full yet, generally very few samples are missing. Then the probability that it is still not full after filling from the second generation is very low and using a third generation is almost never necessary.

\subsection{Comparison to deep reinforcement learning}

Deep reinforcement learning (\drl) approaches are considered more sample efficient than evolutionary approaches \citep{sigaud2018policy}. Being step-based, as opposed to episode-based evolutionary algorithms, \drl methods make a better use of their sample information. In this study, we use \ddpg, an off-policy and sample efficient deep RL algorithm which uses a replay buffer to reuse samples and improve stability \citep{lillicrap2015continuous}. Thus we train a \ddpg agent and compare its sample efficiency with an evolutionary algorithm using the importance mixing mechanism.
\ddpg is designed to solve continuous action RL problems, so for this experiment we work on a continuous version of the CartPole environment. The original CartPole only accepts forces of -1 and +1. Here we relax this constraint by allowing a continuous-valued force in $[-1,1]$.

We choose to use the \snes + EIM algorithm for this study, as Figures~\ref{fig:cp} and \ref{fig:acro} show that it converged with a competitive number of samples with respect to \cmaes + EIM. Besides, using a diagonal covariance matrix makes the training faster, whilst allowing for more expressiveness than the simple \openes algorithm. To limit the computational cost, we performed a limited grid-search over the population size, the initial covariance, and the learning rate of the Adam optimizer. 

\begin{figure}[!ht]
  \centering
     \includegraphics[width=0.6\linewidth]{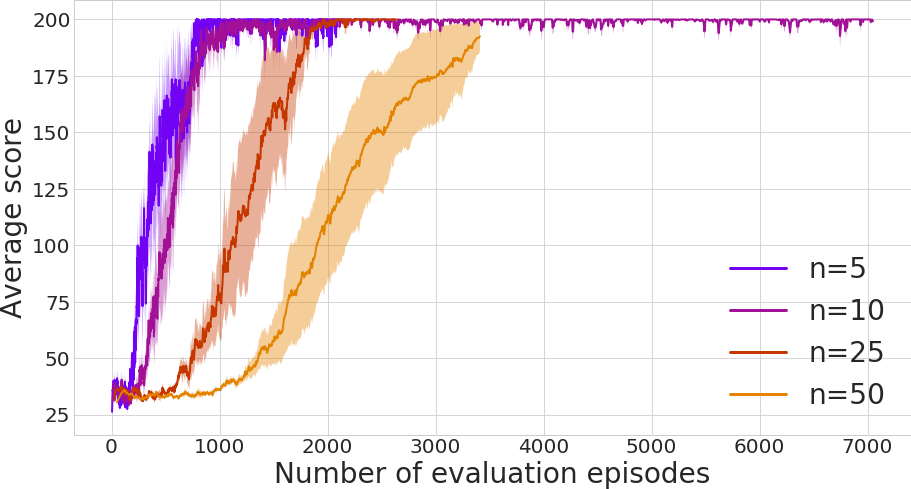}
   \caption{Average score of the population using \snes + EIM as the number of sample evaluations increases for different population sizes. The learning rate is $0.01$ and the initial covariance is $0.1^2\mathcal{I}_d$.}
   \label{fig:pop_size}
\end{figure}

\begin{figure}[!ht]
  \centering
     \includegraphics[width=0.6\linewidth]{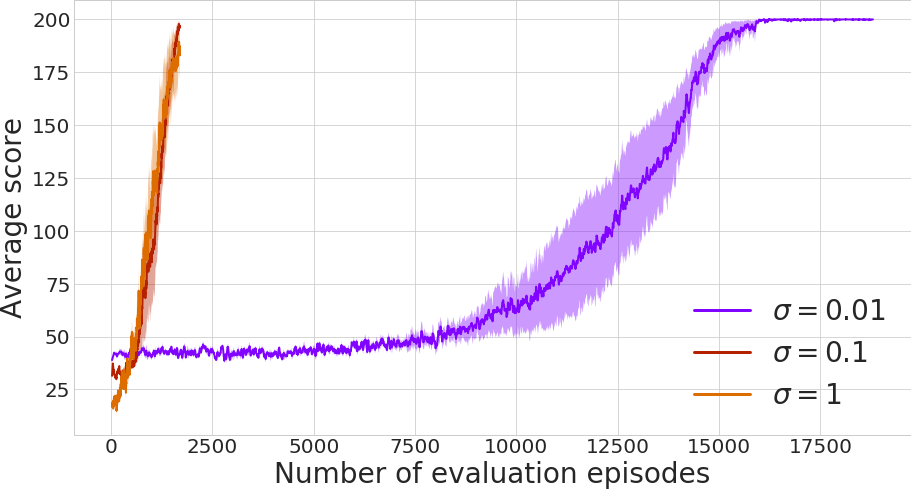}
   \caption{Average score of the population using \snes + EIM as the number of sample evaluations increases for different values of $\sigma$. The learning rate is $0.01$ and the population size is 25.}
   \label{fig:sigma}
\end{figure}

\begin{figure}[!ht]
  \centering
     \includegraphics[width=0.6\linewidth]{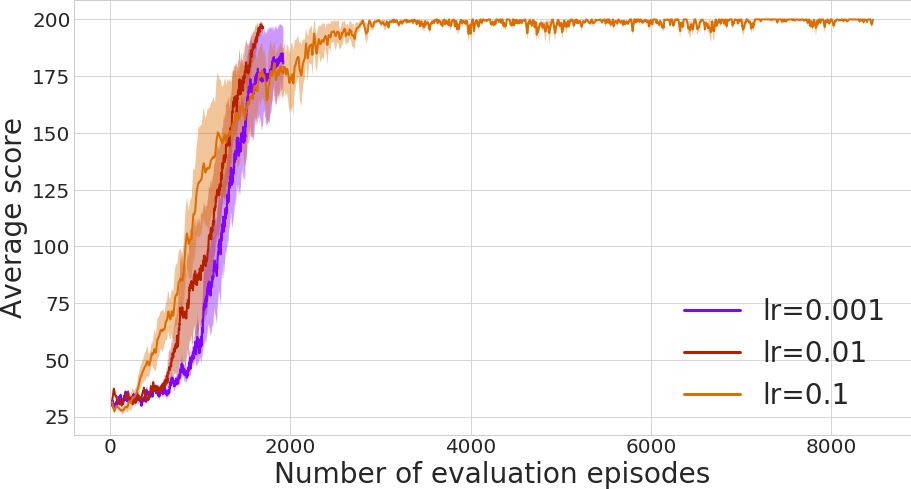}
   \caption{Average score of the population using \snes + EIM as the number of sample evaluations increases for different values of learning rate (lr). The population size is 25 and the initial covariance is $0.1^2\mathcal{I}_d$.}
   \label{fig:learning_rate}
\end{figure}

The influence of the population size can be seen in \figurename~\ref{fig:pop_size}, the influence of $\sigma$ in \figurename~\ref{fig:sigma}, and the influence of the learning rate in \figurename~\ref{fig:learning_rate}. Selecting $popsize=10, \sigma=0.1$ and $lr=0.1$ gives the best results (in terms of convergence speed and stability), with the algorithm converging after around 1500 evaluations.

Then we trained a \ddpg agent \footnote{We used an open source implementation available at \url{https://github.com/liampetti/DDPG}.} on the Continuous CartPole environment, using standard \ddpg parameters as provided in the baselines \citep{baselines}.
Besides, to allow for a fair comparison, the actor had the same architecture as the evolved policy (two hidden layers of size 8). However, we found that we had to increase the size of the hidden layers of the critic to 16 in order for the algorithm to converge. The learning rate of the actor and the critic were respectively set to $0.005$ and $0.01$. The soft-target update parameter was set to $\tau=0.001$, and the discount rate to $\gamma=0.99$. Exploration was ensured with an Ornstein-Uhlenbeck process \citep{PhysRev.36.823}, as proposed in the original \ddpg paper \citep{lillicrap2015continuous}.

\begin{figure}[!ht]
  \centering
  \subfloat[\label{fig:DDPGvSNES}]{\includegraphics[width=0.49\linewidth]{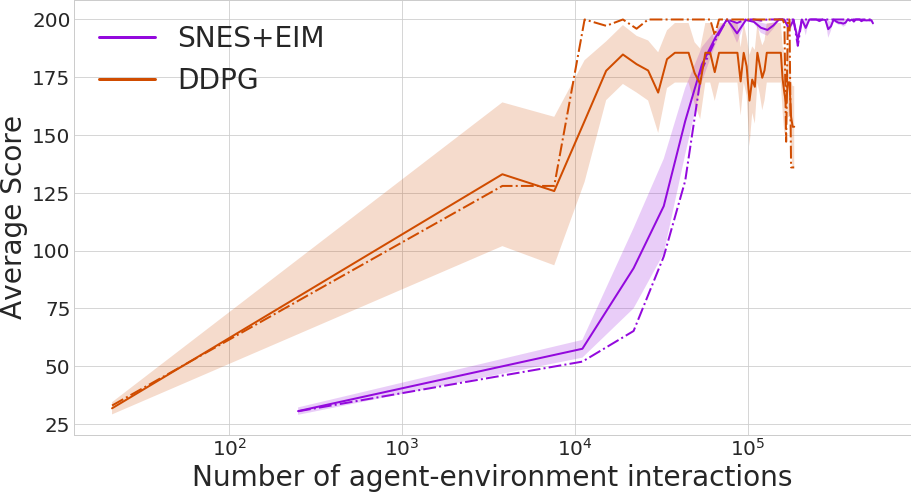}}
     \subfloat[\label{fig:DDPGvSNESHard}]{\includegraphics[width=0.49\linewidth]{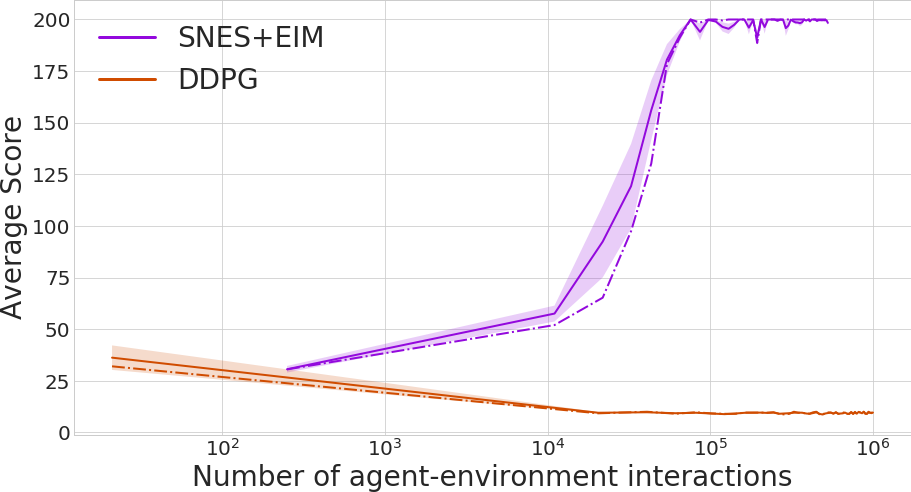}}
   \caption{Average score of the population as the number of sample evaluations increases on the ContinuousCartPole (a) and ContinuousCartPoleHard (b) environments. Dotted lines correspond to medians. For better visualization, we use a log scale over the x-axis.}
\end{figure}

The \snes + EIM algorithm was run for 1500 generations whereas \ddpg was run for 1000 episodes, which is enough in  both cases to reach convergence. The evolution of the performance of \snes + EIM and \ddpg over the number of steps sampled from the system can be seen in \figurename~\ref{fig:DDPGvSNES}. One can see that \ddpg converges around 10 times faster than \snes + EIM on this benchmark. Even with a sample reuse rate of around $90\%$, the gap between evolutionary algorithms and \drl is still important. However, \ddpg is somehow more unstable than \snes + EIM. It sometimes happens that the algorithm diverges late in the experiment, in contrast to the evolutionary method.

However, the Continuous Cartpole benchmark is favorable to \drl methods, since the reward signal is very dense (one reward per step). By contrast, evolutionary algorithms really shine when rewards are sparse, since they only take into account rewards over episodes when enhancing their policies. 
We thus finally compared \ddpg and \snes + EIM on another environment that we call "ContinuousCartPoleHard". In this environment, the agent only receives the total reward after the episode ends. We used the same hyper-parameters for both algorithms, and let \snes + EIM run for 1500 generations, and \ddpg for 1 million time steps.
The change had no influence over the evolutionary algorithms, while it turned out to be much more challenging for the \ddpg agent, as can be seen in \figurename~\ref{fig:DDPGvSNESHard} where \ddpg does not make any progress.

  %%%%%%%%%%%%%%%%%%%%%%%%%%%%%%%%%%%%%%%%%%%%%%%%%%%%%%%%%%%%%%%%%%%%%%%%%%%%%%%%%%
%  \section{Related work}
%  \label{sec:related}

 %%%%%%%%%%%%%%%%%%%%%%%%%%%%%%%%%%%%%%%%%%%%%%%%%%%%%%%%%%%%%%%%%%%%%%%%%%%%%%%%%%
  \section{Conclusion and future work}
  \label{sec:conclu}

In this paper we have given a didactic presentation of an improved version of the importance mixing mechanism initially published in \cite{sun2009efficient}, and we have explained how it could be extended to make profit of the samples coming from several generations.

The experimental study revealed that, on one side, the basic importance mixing mechanism provides significant sample efficiency improvement but, on the other side, the extended importance mixing mechanism does not bring this improvement much further. Besides, importance mixing alone is not enough for evolutionary strategies to compete with deep RL methods in terms of sample efficiency. However, as already noted several times in the literature, evolutionary methods are generally more stable than deep RL methods.

Based on these conclusions, in future work we intend to turn towards frameworks where evolutionary methods and deep RL methods are combined, such as \cite{khadka2018evolutionary} and \cite{colas2018gep}. In both cases, the evolutionary part is not based on an evolutionary strategy, thus the importance mixing mechanism cannot be applied straightforwardly to these combinations. This will be the focus of our next research effort.

%  \bibliography{local}
\bibliographystyle{elsarticle-harv}

%\appendix
%\section{Learning curves}

%\bibliography{/home/sigaud/Bureau/sigaud/Latex/Biblio/motor_learning,/home/sigaud/Bureau/sigaud/Latex/Biblio/rl,/home/sigaud/Bureau/sigaud/Latex/Biblio/perso,/home/sigaud/Bureau/sigaud/Latex/Biblio/continuous_rl,/home/sigaud/Bureau/sigaud/Latex/Biblio/robot_learning,/home/sigaud/Bureau/sigaud/Latex/Biblio/deep,/home/sigaud/Bureau/sigaud/Latex/Biblio/philo,/home/sigaud/Bureau/sigaud/Latex/Biblio/mabiblio,/home/sigaud/Bureau/sigaud/Latex/Biblio/curiosity}

\end{document}